\documentclass[11pt,a4paper]{article}
\usepackage[hyperref]{acl2017}
\usepackage{times,algorithmic,algorithm,amsfonts,amsmath}
\usepackage{url}
\usepackage{latexsym}
\usepackage{booktabs,multirow}
\usepackage{footmisc}

\usepackage{xcolor}
\definecolor{nice-red}{HTML}{E41A1C}
\definecolor{nice-orange}{HTML}{FF7F00}
\definecolor{nice-yellow}{HTML}{FFC020}
\definecolor{nice-green}{HTML}{4DAF4A}
\definecolor{nice-blue}{HTML}{377EB8}
\definecolor{nice-purple}{HTML}{984EA3}

\aclfinalcopy % Uncomment this line for the final submission
\def\aclpaperid{3331} %  Enter the acl Paper ID here

\title{Multi-Task Learning of Keyphrase Boundary Classification}

\author{Isabelle Augenstein \footnotemark \\
  Department of Computer Science \\
  University College London \\
  {\tt i.augenstein@ucl.ac.uk} \\\And
  Anders S{\o}gaard \footnotemark[1] \\
  Department of Computer Science \\
  University of Copenhagen \\
  {\tt soegaard@di.ku.dk} \\
  }

\date{}

\begin{document}
\maketitle
\begin{abstract}
Keyphrase boundary classification (KBC) is the task of detecting keyphrases in scientific articles and labelling them with respect to predefined types. Although important in practice, this task is so far underexplored, partly due to the lack of labelled data. 
To overcome this, we explore several auxiliary tasks, including semantic super-sense tagging and identification of multi-word expressions, and cast the task as a multi-task learning problem with deep recurrent neural networks. Our multi-task models perform significantly better than previous state of the art approaches on two scientific KBC datasets, particularly for long keyphrases. 
%We present a novel approach to keyphrase boundary classification (KBC) based on multi-task recurrent neural networks. KBC is the task of detecting keyphrases in scientific articles and labeling them with respect to predefined concept classes. We explore several auxiliary tasks, including semantic supersense tagging and identification of multi-word expressions. Our multi-task models perform significantly better than previous state of the art approaches on two scientific KBC datasets, with error reductions of up to 14.9\%~on the SemEval 2017 Task 10 corpus and of up to 5.1\%~on the ACL RD-TEC corpus. 
\end{abstract}

\newenvironment{starfootnotes}
  {\par\edef\savedfootnotenumber{\number\value{footnote}}
   \renewcommand{\thefootnote}{$\star$} 
   \setcounter{footnote}{0}}
  {\par\setcounter{footnote}{\savedfootnotenumber}}

\begin{starfootnotes}
\footnotetext{Both authors contributed equally}
\end{starfootnotes}

\section{Introduction}

% Task itself is novel-ish - not introduced in this paper per se, but other papers don't introduce methods for it. Yet.
% We have a small data problem here which we overcome with studying which auxiliary tasks are useful. We do this with a data analysis and testing several additional datasets.
% In doing so, we outperform other methods which are jsut straight-forward sequence labelling methods

The scientific keyphrase boundary classification (KBC) task consists of a) determining keyphrase boundaries, and b) labelling keyphrases with their types according to a predefined schema.
KBC is motivated by the need to efficiently search scientific literature, which can be summarised by their keyphrases. Several companies are working on keyphrase-based recommender systems for scientific literature or search interfaces where scientific articles decorate graphs, in which nodes are keyphrases. Such keyphrases must be dynamically retrieved from the articles, because important scientific concepts emerge on a daily basis, and the most recent concepts are typically the ones of interest to scientists.

KBC is not a common task in NLP, and there are only few small annotated datasets for inducing supervised KBC models, made available recently \cite{conf/lrec/QasemiZadehS16,augenstein-EtAl:2017:SemEval}. Typical KBC approaches therefore rely on hand-crafted gazetteers \cite{hasan-ng:2014:P14-1} or reduce the task to extracting a list of keyphrases for each document \cite{kim-EtAl:2010:SemEval} instead of identifying mentions of keyphrases in sentences.
For related more common NLP tasks such as named entity recognition and identification of multi-word expressions, neural sequence labelling methods have been shown to be useful \cite{lample-EtAl:2016:N16-1}.
In order to overcome the small data problem, we study using more widely available data for tasks related to KBC and exploit their synergies in a deep multi-task learning setup. 

Multi-task learning has become popular within natural language processing and machine learning over the last few years; in particular, {\em hard} parameter sharing of hidden layers in deep learning models. This approach to multi-task learning has three advantages: a) It significantly reduces Rademacher complexity \cite{Baxter:00,Maurer:07}, i.e., the risk of over-fitting, b) it is space-efficient, reducing the number of parameters, and c) it is easy to implement. 

This paper shows how hard parameter sharing can be used to improve gazetteer-free keyphrase boundary classification models, by exploiting different syntactically and semantically annotated corpora, as well as more readily available data such as hyperlinks. %We emphasise the fact that our models are gazetteer-free, making them more appropriate for dynamic keyphrase boundary classification. 

\paragraph{Contributions} We study the so far widely underexplored, though in practice important task of scientific keyphrase boundary classification, for which only a small amount of training data is available. We overcome this by identifying good auxiliary tasks and cast it as a multi-task learning problem. We evaluate our models across two new, manually annotated corpora of scientific articles and outperform single-task approaches by up to 9.64\% F1, mostly due to better performance for long keyphrases.

%\paragraph{Contributions} Our contributions are as follows: We frame keyphrase boundary classification as a multi-task learning problem and identify good auxiliary tasks. We present a series of multi-task learning models for keyphrase boundary classification, induced from mixtures of labeled data and such auxiliary sources. We evaluate these models across two new, manually annotated corpora of scientific articles, the ACL RD-TEC 2.0 corpus and the SemEval 2017 Task 10 dataset.

%\paragraph{Contributions} Our contributions are as follows: We present a series of multi-task learning models for keyphrase boundary classification, induced from mixtures of labeled data and auxiliary sources. We evaluate these models across two new, manually annotated corpora of scientific articles, the ACL RD-TEC 2.0 corpus and the SemEval 2017 Task 10 dataset.

%\ldots

\section{Keyphrase Boundary Classification}

Consider the following sentence from a scientific paper: 

\begin{itemize}
\item[(1)] We find that simple interpolation methods, like log-linear and linear interpolation, improve the performance but fall short of the performance of an oracle. 
\end{itemize}

This sentence occurs in the ACL RD-TEC 2.0 corpus. Here, {\em interpolation methods}~and {\em log-linear and linear interpolation} are annotated as technical keyphrases, {\em performance}~as a keyphrase related to measurements, and {\em oracle}~is a keyphrase labelled as miscellaneous. %The ACL RD-TEC 2.0 corpus defines seven such keyphrase types; the SemEval 2017 Task 10 defines three. %See Table~\ref{tab:DataStatistics} for more dataset characteristics. 
Below, we are interested in predicting the boundaries {\em and}~the types of all keyphrases. %Our evaluation metric is the token micro-averaged $F_1$-score. 

%\ia{Explain input, output labels, evaluation metrics. What tagging scheme do we use? IO, right?}
%\ia{We can move the not-so-exciting parts of this to later parts of the paper before submitting}
%We use the IO tagging scheme due to only using a small corpus. Possible tags are Material, Process, Task. For evaluation, all concurrent tokens of with the same predicted tag are considered to belong to the same mention.
%Although the SemEval 2017 is annotated for multi-label classification (i.e. every token can have multiple labels), we simplify the corpus by reducing it to a single possible label per token.

\section{Multi-Task Learning}\label{sec:Multitask}

Multi-task learning is an approach to learning, in which generalisation is improved by taking advantage of the inductive bias in training signals of related tasks. %Multi-task learning can be motivated in different ways. 
When abundant labelled data is available for an auxiliary task, but little data for the target task, multi-task learning can act as a form of semi-supervised learning combined with a distant supervision signal. Inducing a model from only the sparse target task data may lead to overfitting to random noise in the data, but relying on auxiliary data helps the model generalise, making it easier to abstract away from noise, as well as leveraging the marginal distribution of auxiliary input data. From a representation learning perspective, auxiliary tasks can be used to induce representations that may be beneficial for the target task. %This perspective is particularly important for our analysis below. 
\newcite{Caruana:93} also suggests that the auxiliary task can help focus attention in the induction of the target task model. Finally, multi-task learning can be cast as a regulariser as studies show reductions in Rademacher complexity in multi-task architectures over single-task architectures \cite{Baxter:00,Maurer:07}.
 
Here, we follow the probably most common approach to multi-task learning, known as {\em hard parameter sharing}. This was introduced in \newcite{Caruana:93} in the context of deep neural networks, in which hidden layers can be shared among tasks.
We assume $T$ different training set, $D_1,\cdots,D_T$, where each $D_t$ contains pairs of input-output sequences $(w_{1:n}, y^t_{1:n})$, $w_i \in V$, $y^t_i \in L^t$.  The input vocabulary $V$ is shared across tasks, but the output vocabularies (tagset) $L^t$ are task dependent. At each step in the training process we choose a random task $t$, followed by a random training instance $(w_{1:n}, y^t_{1:n}) \in D_t$.
We use the tagger to predict the labels $\hat{y}^t_i$, suffer a loss with
respect to the true labels $y^t_i$ and update the model parameters. The parameters are trained jointly for a sentence, i.e. cross-entropy loss over each sentence is employed. Each task is associated with an independent classification function, but all tasks share the hidden layers. Note that for our experiments, we only consider one auxiliary task at a time.

%In {\em soft parameter sharing}, each task has separate parameters and separate hidden layers, but the loss at the outer layer is regularized by the current distance between the models. In \newcite{Duong:ea:15}, for example, the loss is regularized by the $L_2$ distance between (selective parts of) the main and auxiliary models. 

%While hard parameter sharing is easy to implement and significantly reduces Rademacher complexity \cite{Baxter:00,Maurer:07}, this is not necessarily the case for soft parameter sharing. Soft parameter sharing is thus, depending on the weight of the regularization term, more prone to overfitting. On the other hand, it gives the model some wiggle room, potentially making multi-task learning applicable under more circumstances, and enabling us to leverage larger volumes of training data when available. So, while we limit ourselves to hard parameter sharing here, we cannot rule out that some form of soft parameter sharing will not fair (even) better.  

\setlength{\tabcolsep}{0.3em}
\begin{table*}[t]
\fontsize{10}{10}\selectfont
\begin{center}
\begin{tabular}{l l l}
\toprule
& \bf SemEval 2017 Task 10 & \bf ACL RD-TEC \\
\midrule
 Labels & Material, Process, Task & Technology and Method,  \\
       &  & Tool and Library, \\
       &  & Language Resource, \\
       &  & Language Resource Product, \\
       &  & Measures and Measurements, \\
       &  & Models, Other \\
Topics & Computer Science, Physics, & Natural Language Processing  \\
 & Material Science &  \\
Number all keyphrases & 5730 & 2939 \\
%Number unique keyphrases & 1697 & 2132 \\
Proportion singleton keyphrases & 31\% & 83\% \\
Proportion single-word mentions & 18\% & 23\% \\
Proportion mentions with word length $>=$ 2 & 82\% & 77\% \\
Proportion mentions with word length $>=$ 3 & 51\% & 33\% \\
Proportion mentions with word length $>=$ 5 & 22\% & 8\% \\
%Proportion mentions are noun phrases & 93\% & 95\% \\
%Most common keyphrases & `Isogeometric analysis', & `model', `features',\\
% & `samples', `calibration process' & `sentences', `sentence', \\
% & `Zirconium alloys', `B decays'  & `words' \\
\bottomrule
\end{tabular}
\end{center}
\caption{\label{tab:DataStatistics} Characteristics of SemEval 2017 Task 10 and ACL-RD-TEC corpora, statistics of training sets}
\end{table*}

\section{Experiments}

\paragraph{Experimental Setup}
We perform experiments for both keyphrase boundary identification ({\em unlabelled}), and keyphrase boundary identification and classification ({\em labelled}). Metrics measured are token-level precision, recall and F1, which are micro-average results across keyphrase types. Types are defined by the two datasets studied.

\paragraph{Auxiliary tasks} We experiment with five auxiliary tasks: (1) syntactic chunking using annotations extracted from the English Penn Treebank, following \newcite{Soegaard:Goldberg:16}; (2) frame target annotations from FrameNet 1.5 (corresponding to the target identification and classification tasks in \newcite{Das:ea:14}); (3) hyperlink prediction using the dataset from \newcite{spitkovsky2010}, (4) identification of multi-word expressions using the Streusle corpus \cite{schneider2015corpus}; and (5) semantic super-sense tagging using the Semcor dataset, following \newcite{Johannsen2014}. We train our models on the main task with one auxiliary task at a time. Note that the datasets for the auxiliary tasks are not annotated with keyphrase boundary identification or classification labels.

\paragraph{Datasets}
We evaluate on the SemEval 2017 Task 10 dataset \cite{augenstein-EtAl:2017:SemEval} and the the ACL RD-TEC 2.0 dataset \cite{conf/lrec/QasemiZadehS16}. The SemEval 2017 dataset is annotated with three keyphrase types, the ACL RD-TEC dataset with seven. For the former, we test on the development portion of the dataset, as the test set is not released yet. We randomly split ACL RD-TEC into a training and test set, reserving 1/3 for testing.
Key dataset characteristics are summarised in Table \ref{tab:DataStatistics}. One important observation is that the SemEval 2017 dataset contains a significantly higher proportion of long keyphrases than the ACL dataset. %Also, ACL RD-TEC contains a large proportion of keyphrases which only appear once in the training set (singletons).

\paragraph{Models} Our single- and multi-task networks are three-layer, bi-directional LSTMs \cite{graves2005framewise} with pre-trained {\sc Senna} embeddings.\footnote{\url{http://ronan.collobert.com/senna/}} For the multi-task networks, we follow the training procedure outlined in Section \ref{sec:Multitask}. The dimensionality of the embeddings is 50, and we follow \newcite{Soegaard:Goldberg:16} in using the same dimensionality for the hidden layers. We add a dropout of 0.1 to the input and train these architectures with momentum SGD with initial learning rate of 0.001 and momentum of 0.9 for 10 epochs. %See the code for details. 

\paragraph{Baselines} Our baselines are \newcite{Finkel:ea:05}\footnote{\url{http://nlp.stanford.edu/software/} \\ \-\hspace{.75cm}  \url{CRF-NER.shtml}} and \newcite{lample-EtAl:2016:N16-1}\footnote{\url{https://github.com/clab/} \\ \-\hspace{.75cm}  \url{stack-lstm-ner}}, in order to compare to a lexicalised and a state-of-the-art neural method. We use the implementations released by the authors and re-train models on our data.

\setlength{\tabcolsep}{0.3em}
\begin{table*}[t]
\fontsize{10}{10}\selectfont
\begin{center}
\begin{tabular}{l c c c | c c c}
\toprule
&\multicolumn{3}{c}{\bf Unlabelled}&\multicolumn{3}{c}{\bf Labelled}\\
\midrule
\bf Method & \bf Precision & \bf Recall & \bf F1 & \bf Precision & \bf Recall & \bf F1 \\
\midrule
\newcite{Finkel:ea:05} & 77.89 & 50.27 & 61.10 & 49.90 & 27.97 & 35.85 \\
\newcite{lample-EtAl:2016:N16-1} & 71.92 & 49.37 & 58.55 & 41.36 & 28.47 & 33.72 \\
\midrule
BiLSTM  & 81.58&57.86&67.71 &45.80&32.48&38.01 \\
\midrule
BiLSTM + Chunking &  82.88&52.08&63.96 & 55.54& 34.90& 42.86\\
BiLSTM + Framenet & 77.86&56.05 &65.18& 54.04& 38.91& 45.24\\
BiLSTM + Hyperlinks &  76.59&60.53&67.62 & 46.99& 44.09& 41.13\\
BiLSTM + Multi-word & 74.80&70.18&{\bf 72.42}  & 46.99& 44.09& {\bf 45.49}\\
BiLSTM + Super-sense &  83.70&51.76 &63.93& 56.94& 35.25& 43.54\\

%LSTM + FN Targets & 86.11 & 69.93 & 79.27 & {\bf 74.31} \\
\bottomrule
\end{tabular}
\end{center}
\caption{\label{tab:ResultsDetectionSemEval} Results for keyphrase boundary classification on the SemEval 2017 Task 10 corpus}
\end{table*}

%\setlength{\tabcolsep}{0.3em}
%\begin{table*}[t]
%\fontsize{10}{10}\selectfont
%\begin{center}
%\begin{tabular}{l c c c c}
%\toprule
%\bf Method & \bf Accuracy & \bf Precision & \bf Recall & \bf F1 \\
%\hline
%\newcite{Finkel:ea:05} & 90.79 &  84.16 & 80.08 & 82.07 \\

%\bottomrule
%\end{tabular}
%\end{center}
%\caption{\label{tab:ResultsDetectionACL} Results for keyphrase boundary identification on ACL corpus}
%\end{table*}

\setlength{\tabcolsep}{0.3em}
\begin{table*}[t]
\fontsize{10}{10}\selectfont
\begin{center}
\begin{tabular}{l  c c c | ccc }
\toprule
&\multicolumn{3}{c}{\bf Unlabelled}&\multicolumn{3}{c}{\bf Labelled}\\
\midrule
\bf Method & \bf Precision & \bf Recall & \bf F1 & \bf Precision & \bf Recall & \bf F1 \\
\midrule
\newcite{Finkel:ea:05} & 84.16 & 80.08 & {\bf 82.07} & 59.97 & 53.86 & 56.75 \\
\newcite{lample-EtAl:2016:N16-1} & 65.60 & 86.06 & 74.45 & 31.30 & 41.07 & 35.53 \\
\midrule
BiLSTM  &  83.40 & 80.36 &  81.85 &59.62&57.45&58.51\\
\midrule
BiLSTM + Chunking  &83.36&79.46& 81.37  &59.26&57.24&57.84 \\
BiLSTM + Framenet & 84.11&79.39&81.68 &60.64&57.24&{58.89}\\
BiLSTM + Hyperlinks & 83.94&79.12&81.46&60.18&56.73&{58.40}\\
BiLSTM + Multi-word & 84.86&76.92&80.69&59.81&54.21&{56.87}\\
BiLSTM + Super-sense & 84.67&78.29&81.36&61.35&56.73&{\bf 58.95}\\
\bottomrule
\end{tabular}
\end{center}
\caption{\label{tab:ResultsDetectionACL} Results for keyphrase boundary classification on the ACL RD-TEC corpus}
\end{table*}

\section{Results and Analysis}

Results for SemEval 2017 Task 10 corpus are presented in Table~\ref{tab:ResultsDetectionSemEval}, and for the ACL RD-TEC corpus in Table~\ref{tab:ResultsDetectionACL}. For the SemEval corpus, all five labelled multi-task learning models outperform both examples of previous work, as well as our single-task BiLSTM baseline, by some margin. For ACL RD-TEC, three of out five multi-task learning labelled labelled perform better than the single-task BiLSTM baseline. %For the unlabelled case, most multi-task models are on par with the LSTM baseline. %and an error reduction of 2.2\% is achieved.

On the SemEval corpus, the F1 error reduction of of the best labelled model over the Stanford tagger is 9.64\%. The lexicalised \newcite{Finkel:ea:05} model shows a surprisingly competitive performance on the ACL RD-TEC corpus, where it is only 2 points in F1 behind our best performing labelled model and on par with our best-performing unlabelled model. Results with \newcite{lample-EtAl:2016:N16-1}, on the other hand, are lower than the \newcite{Finkel:ea:05} baseline. This might be due to the model having a large set of parameters to model state transitions which poses a difficulty for small training datasets.

Overall, multi-task models show bigger improvements over baselines for the SemEval corpus, and all models achieve better results on ACL RD-TEC. Statistics shown in Table \ref{tab:DataStatistics} help to explain this. Most noticeably, the SemEval dataset contains a significantly higher proportion of long keyphrases than the ACL dataset. Interestingly, ACL RD-TEC contains a large proportion of keyphrases which only appear once in the training set (singletons), significantly fewer keyphrases and more keyphrase type, but that does not seem to impact results as much as a high proportion of long keyphrases.

%ACL RD-TEC contains a large proportion of keyphrases which only appear once in the training set (singletons), but that does not seem to have a noticeable effect on the results.

%\paragraph{Datasets}
%Comparing the results for the two datasets, we observe that the results for the SemEval 2017 Task 10 corpus are consistently lower than for the  ACL RD-TEC corpus. In order to understand why we observe this, we compare statistics about the datasets (computed based on the training data subsets), summarised in Table \ref{tab:DataStatistics}.\\

%The main difference noticeable between the two datasets that might indicate why one is more difficult to train a model for than the other is that the SemEval 2017 Task 10 dataset contains a significantly higher proportion of long keyphrases than the ACL RD-TEC dataset. ACL RD-TEC contains a large proportion of keyphrases which only appear once in the training set (singletons), but that does not seem to have a noticeable effect on the results.

%We further investigate what common keyphrases models fail to recognise. 
All models struggle with semantically vague or broad keyphrases (e.g. `items', `scope', `key') and long keyphrases, especially those containing clauses (e.g. `complete characterisation of the oxide particles', `earley deduction proof procedure for definite clauses'). The multi-task models generally outperform the BiLSTM baseline for long phrases (e.g. `language-independent system for automatic discovery of text in parallel translation', `honeycomb network of graphite bricks'). %, especially when trained to predict bracketing. 
Being able to recognise long keyphrases correctly is part of the reason our multi-task models outperform the baselines, especially on the SemEval dataset, which contains many such long keyphrases.

\section{Related Work}
\paragraph{Multi-Task Learning}
Hard sharing of all hidden layers was introduced in \newcite{Caruana:93}, and popularised in NLP by \newcite{Collobert:ea:11}. Several variants have been introduced, including hard sharing of selected layers \cite{Soegaard:Goldberg:16} and sharing of parts (subspaces) of layers \cite{Liu:ea:15}. \newcite{Soegaard:Goldberg:16} show that hard parameter sharing is an effective regulariser, also on heterogeneous tasks such as the ones considered here. Hard parameter sharing has been studied for several tasks, including CCG super tagging \cite{Soegaard:Goldberg:16}, text normalisation \cite{Bollman:ea:16}, neural machine translation \cite{dong-EtAl:2015:ACL-IJCNLP2,Luong:ea:16}, and super-sense tagging \cite{Alonso:Plank:16}.
Sharing of information can further be achieved by extending LSTMs with an external memory shared across tasks \cite{liu-qiu-huang:2016:EMNLP2016}.
A further instance of multi-task learning is to optimise a supervised training objective jointly with an unsupervised training objective, as shown in \newcite{yu-buys-blunsom:2016:EMNLP2016} for natural language generation and auto-encoding, and in \newcite{rei2017ACL} for different sequence labelling tasks and language modelling.

\paragraph{Boundary Classification}
KBC is very similar to named entity recognition (NER), though arguably harder. Deep neural networks have been applied to NER in \newcite{Collobert:ea:2011natural,lample-EtAl:2016:N16-1}. Other successful methods rely on conditional random fields, thereby modelling the probability of each output label conditioned on the label at the previous time step. \newcite{lample-EtAl:2016:N16-1}, currently state-of-the-art for NER, stack CRFs on top of recurrent neural networks. We leave exploring such models in combination with multi-task learning for future work. %Early work uses lexical features \cite{Finkel:ea:05}, followed by word embedding features \cite{Turian:ea:10} and most recently biLSTMs \cite{lample-EtAl:2016:N16-1}. The latter method currently holds the state of the art performance on the ConLL 2003 data.

Keyphrase detection methods specific to the scientific domain often use keyphrase gazetteers as features or exploit citation graphs \cite{hasan-ng:2014:P14-1}. However, previous methods relied on corpora annotated for type-level identification, not for mention-level identification \cite{kim-EtAl:2010:SemEval,sterckx-EtAl:2016:EMNLP2016}. While most applications rely on extracting keyphrases (as types), this has the unfortunate consequence that previous work ignores acronyms and other short-hand forms referring to methods, metrics, etc. Further, relying on gazetteers makes overfitting likely, obtaining lower scores on out-of-gazetteer keyphrases.

%Finally, there is some existing work on extracting keyphrases from domains other than scientific publications. \newcite{sterckx-EtAl:2016:EMNLP2016} extract lists of keyphrases from newspapers and magazines using a supervised methods that learns from noisy labels by multiple annotators. They also apply their trained classifiers to scientific articles with author-defined lists of keyphrases. \cite{zhang-EtAl:2016:EMNLP20164}

\section{Conclusions and Future Work}

We present a new state of the art for keyphrase boundary classification, using data from related, auxiliary tasks; in particular, super-sense tagging and identification of multi-word expressions. Deep multi-task learning improves significantly on previous approaches to KBC, with error reductions of up to 9.64\%, mostly due to better identification and labelling of long keyphrases.

%There are many ways this work can be improved, however. Hard parameter sharing is a very popular approach to multi-task learning, but there is a wide range of alternatives that we did not consider here. 
%There are many additional experiments left for future work. We only employ hard parameter sharing for multi-task learning, however there are other possible multi-task learning regimes.
In future work, we want to explore alternative multi-task learning regimes to hard parameter sharing and experiment with additional auxiliary tasks.
The auxiliary tasks considered here are standard NLP tasks, hyperlink prediction aside. Other tasks may be more directly relevant such as predicting the layout of calls for papers for scientific conferences, or predicting hashtags in tweets by scientists, since both data sources contain scientific keyphrases. %We use a greedy recurrent neural network taking sequences of words as input. Better performance can probably be obtained with LSTM-CRFs and character-based encoding. 

%- better multi-task learning (better tasks, combining soft+hard parameter sharing)
%- more in-task data - combining with learning to generate training data \cite{conf/emnlp/MiaoB16,conf/emnlp/BouchardSR16} or distantly supervised annotation with keyphrase gazetteers, as performed in previous work.
%- better neural base model (with CRFs - not clear if that really works from our baselines, but could be down to not enough data; with character RNNs for unseen words)

%It is likely that we would further improve our results by using a hybrid CRF method as in \cite{lample-EtAl:2016:N16-1}, but we leave that for future work.

\section*{Acknowledgments}

We would like to thank Elsevier for supporting this work.

\bibliography{acl2017}
\bibliographystyle{acl_natbib}

\end{document}